\documentclass[10pt,letterpaper]{article}

\usepackage[T1]{fontenc}
\usepackage[utf8]{inputenc}
\usepackage{times}
\usepackage{microtype}
\usepackage[
  letterpaper,
  top=0.72in,
  bottom=0.75in,
  left=0.82in,
  right=0.82in,
  headheight=13pt,
  headsep=10pt
]{geometry}
\usepackage{amsmath,amssymb,bm}
\usepackage{booktabs,array,tabularx,multirow}
\usepackage{graphicx}
\usepackage{caption}
\usepackage{subcaption}
\usepackage[dvipsnames,table]{xcolor}
\usepackage{tikz}
\usetikzlibrary{arrows.meta,fit,matrix,positioning}
\usepackage[numbers,square,sort&compress]{natbib}
\let\cite\citep
\usepackage{url}
\usepackage[
  colorlinks=true,
  linkcolor=blue,
  citecolor=blue,
  urlcolor=blue,
  pdftitle={Benchmarking Peptide--Protein Affinity Prediction Across Peptide and Target Shifts},
  pdfauthor={Jiaxin Tian, Darren An, Jun Li},
  bookmarksopen=true,
  bookmarksnumbered=true
]{hyperref}
\usepackage[nameinlink,noabbrev]{cleveref}
\usepackage{placeins}
\usepackage{flafter}
\usepackage{titlesec}
\usepackage{fancyhdr}

\crefname{figure}{Figure}{Figures}
\crefname{table}{Table}{Tables}
\crefname{equation}{Eq.}{Eqs.}

\graphicspath{{figures/}}
\titleformat{\section}
  {\fontsize{13}{15}\selectfont\bfseries}
  {\thesection.}{0.45em}{}
\titleformat{\subsection}
  {\fontsize{11}{13}\selectfont\bfseries}
  {\thesubsection.}{0.45em}{}
\titleformat{\subsubsection}
  {\normalsize\scshape}
  {\thesubsubsection.}{0.45em}{}
\titlespacing*{\section}{0pt}{1.25em plus 0.3em minus 0.2em}{0.55em}
\titlespacing*{\subsection}{0pt}{1.0em plus 0.2em minus 0.15em}{0.4em}
\titlespacing*{\subsubsection}{0pt}{0.8em plus 0.2em minus 0.1em}{0.3em}

\renewenvironment{abstract}{%
  \begin{center}
  \begin{minipage}{0.86\linewidth}
  {\centering\fontsize{12}{14}\selectfont\bfseries Abstract\par}
  \vspace{0.35em}
  \small\noindent\ignorespaces
}{%
  \end{minipage}
  \end{center}
  \vspace{0.35em}
}

\newcommand{\SPCC}{\mathrm{SPCC}}
\newcommand{\AULC}{\mathrm{AULC}}

\begin{document}
\thispagestyle{fancy}
\vspace*{-0.15in}
\begin{center}
  {\fontsize{17}{20}\selectfont\bfseries
  Benchmarking Peptide--Protein Affinity Prediction\\
  Across Peptide and Target Shifts\par}
  \vspace{0.65em}
  {\fontsize{12}{14}\selectfont
  Jiaxin Tian\textsuperscript{1}, Darren An\textsuperscript{2}, and
  Jun Li\textsuperscript{2,*}\par}
\end{center}
\begingroup
\renewcommand{\thefootnote}{\arabic{footnote}}
\footnotetext[1]{College of Biology, Hunan University, Changsha, Hunan, China}
\footnotetext[2]{Lingang Laboratory, Shanghai, China}
\renewcommand{\thefootnote}{*}
\footnotetext[3]{Corresponding author: \href{mailto:jun.li@lglab.ac.cn}{\textcolor{black}{jun.li@lglab.ac.cn}}}
\endgroup
\vspace{0.55em}

\begin{abstract}
Peptide--protein affinity models are often evaluated with a single data split, obscuring whether they interpolate among measurements for observed targets or generalize across peptide or target shifts. We integrated three sources of quantitative peptide--protein binding data to obtain 11,349 deduplicated pairs and benchmarked ten peptide representations, ESM-2 protein embeddings, and six regressors under peptide-similarity, within-target, and leave-target-out partitions. Across 60 matched representation--regressor configurations, mean test Spearman correlations were 0.462, 0.669, and 0.530, respectively. The top configuration shifted from ECFP-16 count fingerprints with random forest in the first two settings to HELM-BERT with Extra Trees when exact target sequences were excluded. Representation-rank correlations ranged from $-0.042$ to 0.624 across partitions, whereas regressor-rank correlations ranged from 0.771 to 0.943. Learning curves showed that representation differences were largest with limited supervision and narrowed as training data increased. PeptideCLM-2 adaptation and simple element-wise interaction features provided no consistent gain over a frozen encoder and direct concatenation under the tested protocols. These conclusions are specific to a dataset that pools transformed $K_d$, $K_i$, and $IC_{50}$ measurements and to target exclusion at the exact-sequence level. Peptide--protein affinity benchmarks should therefore align data partitions with the intended use and jointly assess the effects of data scale, molecular representation, and downstream learner.
\end{abstract}

\section{Introduction}

Peptides can engage extended protein surfaces with high affinity and selectivity while remaining accessible to chemical synthesis \cite{muttenthaler2021,wang2022}. This makes them attractive for protein--protein interaction targets that are often difficult for small molecules \cite{scott2016,vanwier2025}. Their development is limited by proteolysis, short exposure, poor oral absorption, and low membrane permeability \cite{zheng2025}. Cyclization, backbone modification, terminal capping, and non-canonical residues can mitigate these liabilities, but greatly enlarge the design space \cite{hickey2023,castro2023,deng2026}.

Computational models can reduce experimental screening by prioritizing peptide--target pairs. Peptides may be encoded as sequences, molecular graphs, circular fingerprints, or pretrained embeddings; proteins can likewise be represented by sequence language models. Earlier affinity models used physicochemical sequence kernels, while recent systems address motif--domain binding, interaction classification, and residue-level binding sites with sequence-, structure-, or attention-based architectures \cite{giguere2013,cunningham2020,lei2021,abdin2022pepnn,li2025pepban,yin2024ppi}. Structural resources such as PepBDB and larger activity collections support complementary analyses \cite{wen2019pepbdb,zhu2025}. However, results across these tasks are not directly comparable because their endpoints, negative examples, and held-out entities differ.

\begin{figure}[!t]
  \centering
  \includegraphics[width=0.92\linewidth]{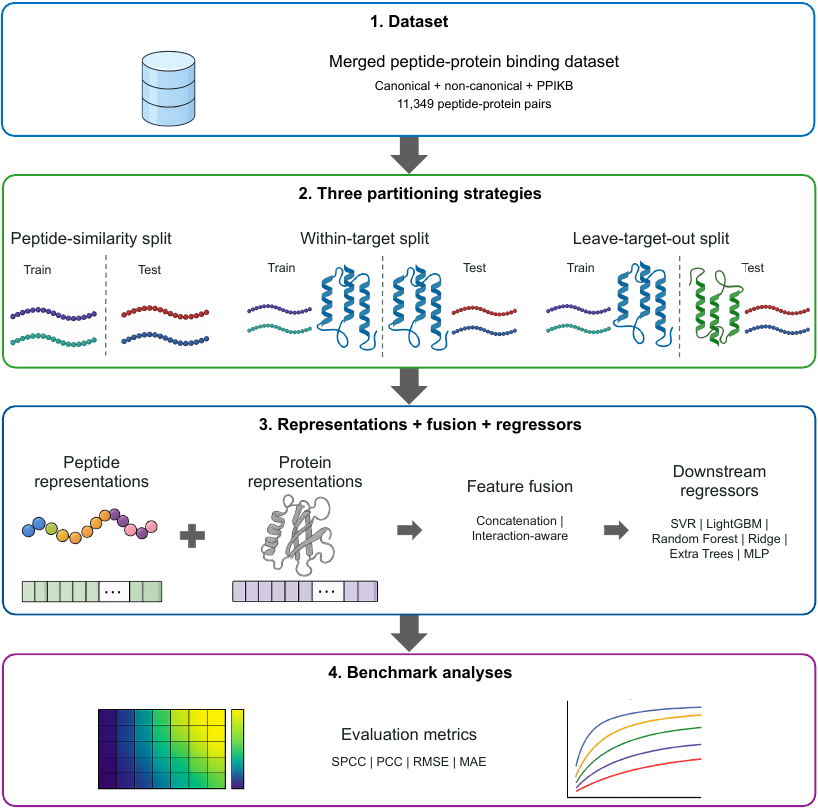}
  \caption{Study design for generalization-aware peptide--protein affinity prediction. Canonical, non-canonical, and PPIKB records were merged and deduplicated at the peptide--target-pair level. Three partitions represented structurally separated peptides, additional measurements for observed targets, and unseen target sequences. Peptide and protein features were generated independently, fused, and supplied to downstream regressors. The benchmark compared generalization regimes, training-set sizes, peptide-encoder adaptation, and pairwise feature-fusion strategies, with SPCC as the primary metric.}
  \label{fig:framework}
\end{figure}

Representation comparisons also depend on data partitioning. Random record-level splits may place close molecular analogues, homologous sequences, or repeated biological entities in both training and test sets, favoring interpolation and potentially overstating prospective performance \cite{sheridan2013,wallach2018,walsh2021,teufel2023,tossou2024}. Fern{\'a}ndez-D{\'i}az et al. showed that representation rankings depended on peptide class and downstream learner, that chemical fingerprints remained competitive, and that transfer between canonical and modified peptides was harder than within-class interpolation \cite{fernandezdiaz2025modified}. These findings show that representation quality cannot be discussed independently of data partition and downstream learner. Pairwise affinity prediction also adds target novelty: a peptide may be chemically distinct while its target is represented in training, or the target sequence itself may be absent. These settings require separate evaluation.

We therefore considered three deployment regimes. Peptide-similarity partitioning tests transfer to structurally separated peptides while allowing targets to recur. Within-target prediction models follow-up measurements for observed proteins. Leave-target-out prediction tests exact protein sequences excluded from fitting. The first two settings can exploit target-specific information learned from other records; the third requires the protein representation and peptide--protein mapping to transfer jointly. It does not, however, guarantee separation between homologous proteins. More generally, molecular out-of-distribution studies show that conclusions can change with the source of distribution shift \cite{fernandezdiaz2025good,tossou2024}.

Our benchmark asks whether performance and model rankings change across these regimes, how training-set size affects the relative performance of peptide representations and downstream learners, and whether adapting a pretrained peptide encoder is more useful than increasing supervision. We also test whether simple element-wise peptide--protein features improve direct concatenation. Three sources of quantitative peptide--protein binding data were integrated and deduplicated to obtain 11,349 measurements for 9,510 peptides and 2,659 target sequences. Ten peptide representations were paired with ESM-2 protein embeddings and six regressors. Learning curves, PeptideCLM-2 adaptation, and feature-fusion experiments then separated the contributions of representation, data scale, and model design. The objective was not a universal winner, but conclusions that remain interpretable across intended uses.

\FloatBarrier
\section{Materials and Methods}

\subsection{Data sources and endpoint transformation}

The benchmark combined three sources of quantitative peptide--protein binding data (\cref{tab:datasets}). Canonical and non-canonical datasets contributed 1,002 and 299 records, respectively \cite{fernandezdiaz2025modified}. The PPIKB preprint reports 21,760 manually validated quantitative affinity entries overall \cite{zhu2025}; the study-specific PPIKB-derived input used here comprised 15,248 canonical records. Records were keyed by peptide SMILES and complete target sequence, then duplicate pairs were removed. The final dataset comprised 11,349 pairs, 9,510 peptides, and 2,659 protein sequences.

\begin{table}[t]
  \centering
  \small
  \caption{Sources and composition of the peptide--protein binding benchmark.}
  \label{tab:datasets}
  \begin{tabularx}{\linewidth}{@{}Xlrl@{}}
    \toprule
    Dataset & Peptide class & Records & Ref. \\
    \midrule
    Protein--peptide binding & Canonical & 1,002 & \cite{fernandezdiaz2025modified} \\
    Protein--peptide binding & Non-canonical & 299 & \cite{fernandezdiaz2025modified} \\
    PPIKB-derived benchmark input\textsuperscript{a} & Canonical & 15,248 & \cite{zhu2025} \\
    Integrated benchmark\textsuperscript{b} & Mixed & 11,349 & \cite{zhu2025,fernandezdiaz2025modified} \\
    \bottomrule
  \end{tabularx}
  \begin{minipage}{\linewidth}
    \footnotesize\textsuperscript{a}Study-specific canonical subset; PPIKB reports 21,760 manually validated quantitative affinity entries overall. \textsuperscript{b}Number remaining after merging the three sources and removing repeated peptide--target pairs using peptide SMILES and target sequence as the pair identifier.
  \end{minipage}
\end{table}

The source records reported dissociation constants ($K_d$), inhibition constants ($K_i$), or half-maximal inhibitory concentrations ($IC_{50}$). Values were first expressed in molar units and then mapped to
\begin{equation}
  S=-\log_{10}(C),
  \label{eq:affinity}
\end{equation}
where $C$ is the reported molar concentration. Larger $S$ indicates stronger apparent affinity. The transformation supplies a common modeling scale without assuming thermodynamic equivalence among endpoints.

\subsection{Molecular representations and pair construction}

Ten peptide featurizations covered different molecular resolutions and pretraining domains (\cref{tab:representations}): 2,048-dimensional binary and count ECFP-16 fingerprints \cite{rogers2010}, PepFuNN \cite{ochoa2025}, chemical language models ChemBERTa-2 and MolFormer-XL \cite{ahmad2022,ross2022}, peptide language models HELM-BERT, PeptideCLM, PeptideCLM-2, and PepTune \cite{lee2025,feller2025,feller2026,tang2025}, and graph-based PepLand \cite{zhang2025pepland}. In the primary representation--regressor benchmark, pretrained neural encoders were frozen to produce fixed-length peptide features, and target proteins were represented by fixed-length ESM-2 features \cite{lin2023}. For each peptide--protein pair, the precomputed peptide and protein vectors were directly concatenated and supplied to each of the six downstream regressors. Task-specific adaptation of PeptideCLM-2 was evaluated separately in \cref{sec:scaling-adaptation}.

\begin{table}[t]
  \centering
  \small
  \caption{Peptide and protein representations included in the benchmark.}
  \label{tab:representations}
  \begin{tabularx}{\linewidth}{@{}lXll@{}}
    \toprule
    Entity & Representation family & Model or feature & Ref. \\
    \midrule
    \multirow{10}{*}{Peptide} & Circular fingerprints & ECFP-16 & \cite{rogers2010} \\
    & & ECFP-16 counts & \cite{rogers2010} \\
    \cmidrule(lr){2-4}
    & Peptide fingerprint & PepFuNN & \cite{ochoa2025} \\
    \cmidrule(lr){2-4}
    & Chemical language models & ChemBERTa-2 & \cite{ahmad2022} \\
    & & MolFormer-XL & \cite{ross2022} \\
    \cmidrule(lr){2-4}
    & Peptide language models & HELM-BERT & \cite{lee2025} \\
    & & PeptideCLM & \cite{feller2025} \\
    & & PeptideCLM-2 & \cite{feller2026} \\
    & & PepTune & \cite{tang2025} \\
    \cmidrule(lr){2-4}
    & Peptide graph model & PepLand & \cite{zhang2025pepland} \\
    \midrule
    Protein & Protein language model & ESM-2 & \cite{lin2023} \\
    \bottomrule
  \end{tabularx}
\end{table}

The separate feature-fusion ablation used an independent neural prediction architecture with frozen PeptideCLM-2 representations and precomputed protein representations. Peptide vector $\bm{p}$ and target vector $\bm{t}$ were mapped to a shared dimensionality through separate learnable projections:
\begin{equation}
  \widetilde{\bm{p}}=f_p(\bm{p}),
  \qquad
  \widetilde{\bm{t}}=f_t(\bm{t}).
  \label{eq:projection}
\end{equation}
Each projection consisted of LayerNorm, a linear mapping, GELU activation, and dropout, and mapped its input to a 512-dimensional latent space. The first fusion strategy concatenated the two projected vectors,
\begin{equation}
  \bm{z}_{\mathrm{concat}}
  =\left[\widetilde{\bm{p}},\widetilde{\bm{t}}\right].
  \label{eq:concat}
\end{equation}
The interaction-aware alternative appended the element-wise product and absolute difference,
\begin{equation}
  \bm{z}_{\mathrm{interaction}}
  =\left[\widetilde{\bm{p}},\widetilde{\bm{t}},
    \widetilde{\bm{p}}\odot\widetilde{\bm{t}},
    \left|\widetilde{\bm{p}}-\widetilde{\bm{t}}\right|\right],
  \label{eq:interaction}
\end{equation}
where $\odot$ denotes element-wise multiplication. Both fusion rules used identical partitions, seeds, projection dimensions, and MLP prediction heads.

\subsection{Generalization-oriented data partitions}

The \emph{peptide-similarity} partition represented structurally separated peptides. Following Hestia-GOOD/CCPart \cite{fernandezdiaz2025good}, similarity used 2,048-bit MAPc fingerprints of diameter 20 (radius 10) and the Jaccard index. Connected-component partitions were generated at thresholds 0.1--1.0 in 0.1 increments. The target test fraction was 20\%; partitions were retained when test records comprised at least 18.5\% of the data, and 10\% of the remaining pool was used for validation. Peptides obeyed the similarity separation, while target sequences could recur. Cross-partition model comparisons and learning curves used threshold 0.5; the full threshold series was used only for the fusion sensitivity analysis.

The \emph{within-target} partition represented follow-up measurements for known targets. Within each exact target sequence, records were assigned to training, validation, and test subsets at approximately 70/10/20. Targets with fewer than three records remained in training. Target sequences, but not individual records, could therefore recur across subsets.

The \emph{leave-target-out} partition grouped all records for an exact protein sequence in one subset. Because target groups were imbalanced, 5,000 randomized assignments were evaluated and the assignment closest to 70/10/20 record proportions was retained (seed 42). Exact sequences could not overlap, but no homology or protein-family threshold was imposed.

\subsection{Downstream learning and model selection}

Six regressors covered linear, kernel, bagged-tree, boosted-tree, and neural approaches: support vector regression (SVR) \cite{cortes1995}, LightGBM \cite{ke2017}, random forest \cite{breiman2001}, ridge regression \cite{hoerl1970}, Extra Trees \cite{geurts2006}, and a multilayer perceptron (MLP) \cite{rumelhart1986}.

Optuna optimized each representation--regressor combination and setting against training-only five-fold cross-validation SPCC \cite{akiba2019}. Searches used 30 trials and stopped after 20 trials without improvement; predefined validation and test subsets were excluded from hyperparameter selection. Selected configurations were refitted on the full training subset and evaluated on the fixed test subset with five random seeds.

\subsection{Training-set scaling and peptide-encoder adaptation}
\label{sec:scaling-adaptation}

To examine the sample efficiency of pretrained peptide representations across supervision levels, learning curves combined five peptide-specific representations---HELM-BERT, PeptideCLM, PeptideCLM-2, PepTune, and PepLand---with all six regressors. Experiments used the fixed peptide-similarity partition at threshold 0.5. Training sizes were 100, 300, 1,000, 3,000, 6,000, and all 8,172 available training records. Validation and test sets were fixed, and optimization was repeated at each size. Model-seed results were first summarized within each training-set subsampling replicate and then averaged across available replicates. Because uncertainty intervals were not retained in the Figure~4 summary, comparisons among these learning curves were interpreted descriptively.

Encoder adaptation used PeptideCLM-2 hybrid-small, distinct from the MLM-small checkpoint in the fixed-representation benchmark. We compared a frozen encoder, unfreezing the final one, two, or four transformer blocks, full fine-tuning, and Low-Rank Adaptation (LoRA) \cite{hu2021}. The frozen strategy trained only the MLP; partial strategies also updated corresponding normalization layers. LoRA adapters were inserted into attention-related linear layers with rank $r=8$, scaling $\alpha=16$, and dropout 0.05.

All strategies used fixed protein features, concatenation, the same MLP head and peptide-similarity partition at threshold 0.5, and all six training sizes. Three matched training-set subsampling replicates were evaluated. Validation loss controlled early stopping and model selection.

\subsection{Evaluation and statistical analysis}

Metrics were Spearman correlation (SPCC), Pearson correlation (PCC), root mean squared error (RMSE), and mean absolute error (MAE); mean squared error was also recorded. SPCC was primary because it measures agreement between the rankings of predicted and experimental values without requiring a linear relationship between them; all comparative analyses reported here use SPCC.

Five-seed results were summarized by mean and standard deviation. Seed-averaged test SPCC for each of the 60 matched representation--regressor combinations was the paired unit in scenario comparisons. Two-sided Wilcoxon signed-rank tests were used with Holm adjustment for three contrasts. This inference quantifies consistency across the evaluated model configurations rather than uncertainty from resampling the underlying dataset. Configurations were ranked by mean SPCC within each scenario, and scenario rank vectors were compared by Spearman correlation. Representation scores were averaged over regressors, regressor scores over representations, and ties received average ranks.

For the adaptation study, each learning curve was summarized by its area under the learning curve (AULC). Training sizes $n_i$ were transformed to a normalized logarithmic coordinate,
\begin{equation}
  x_i=\frac{\log_{10}(n_i)-\log_{10}(n_{\min})}
             {\log_{10}(n_{\max})-\log_{10}(n_{\min})},
  \label{eq:normalized-size}
\end{equation}
and integrated by the trapezoidal rule,
\begin{equation}
  \AULC=\sum_{i=1}^{K-1}
  \frac{y_i+y_{i+1}}{2}\left(x_{i+1}-x_i\right),
  \label{eq:aulc}
\end{equation}
where $y_i$ is SPCC at size $n_i$. All six sizes, including 8,172 records, contributed to AULC; only complete learning curves were analyzed. A Friedman test compared adaptation strategies, using strategy as the repeated condition and each of the three training-set subsampling replicates as a block.

\section{Results}

\subsection{The held-out entity changes both performance and model choice}

\begin{figure}[!htbp]
  \centering
  \includegraphics[width=\linewidth]{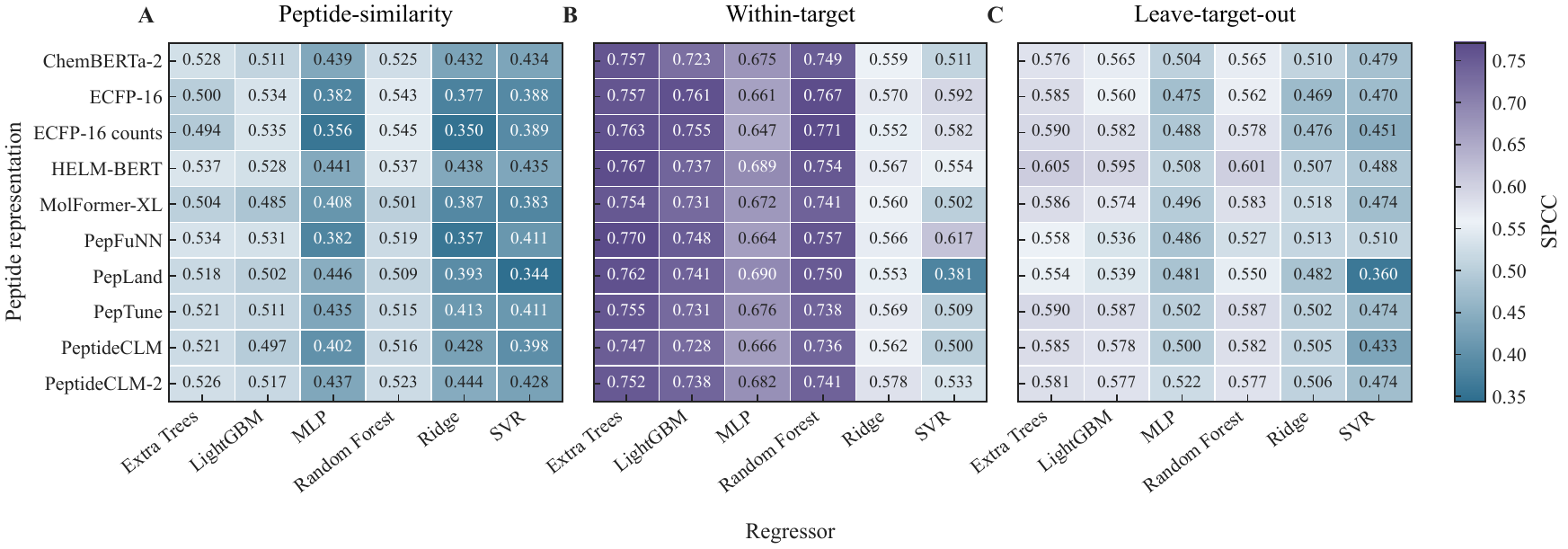}
  \caption{Model-performance landscapes under three definitions of generalization. Each cell reports five-seed mean full-data test SPCC for one peptide representation and downstream regressor under (A) peptide-similarity at threshold 0.5, (B) within-target, or (C) leave-target-out evaluation. The row and column order is identical across panels, and the color scale is shared.}
  \label{fig:landscape}
\end{figure}

\begin{figure}[!htbp]
  \centering
  \includegraphics[width=0.95\linewidth]{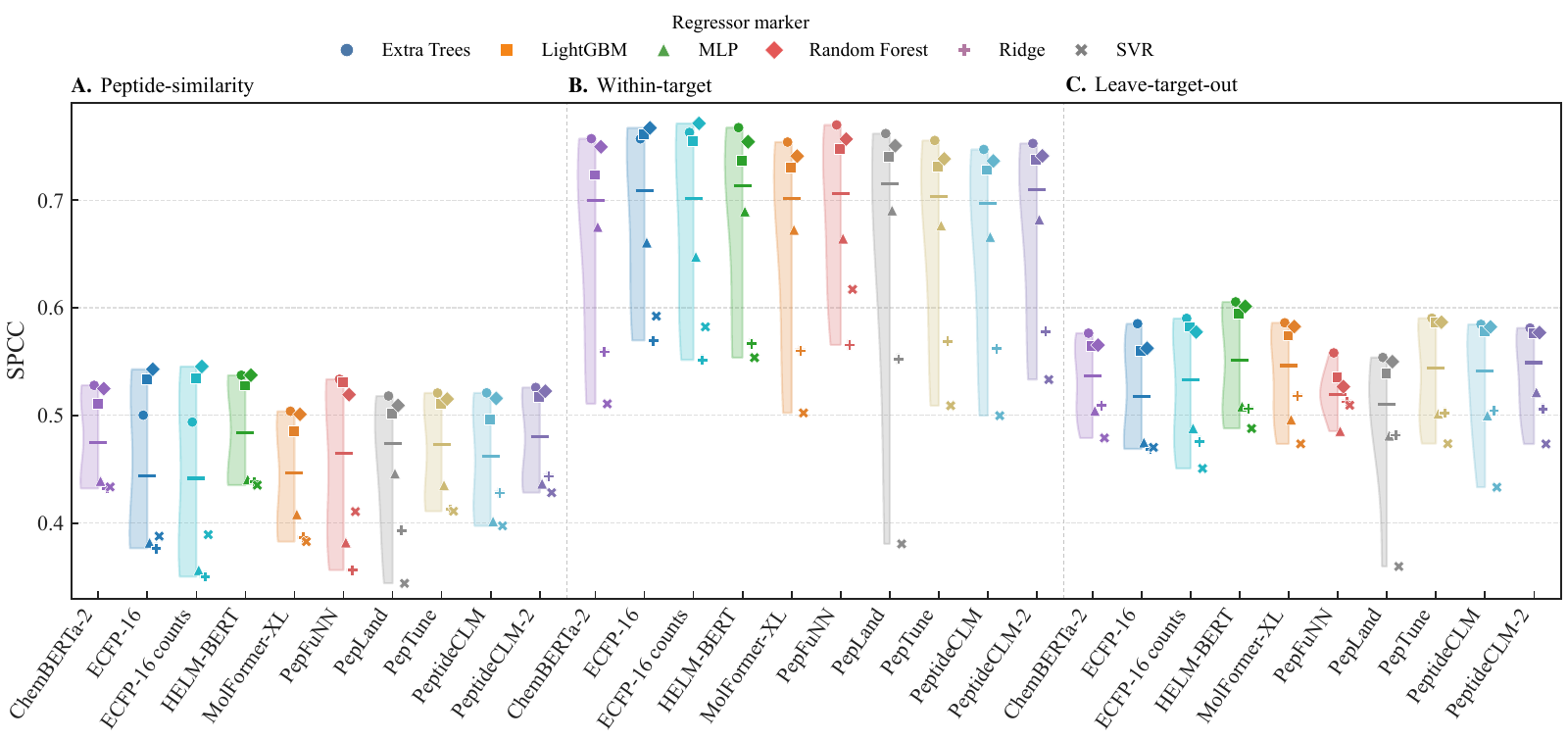}
  \caption{Distribution of full-data test SPCC across peptide representations under (A) peptide-similarity partitioning at threshold 0.5, (B) within-target evaluation, and (C) leave-target-out evaluation. For each representation, the half violin summarizes the six regressor values and the horizontal bar marks their median; the corresponding regressor points are shown alongside. Each marker is the five-seed mean for one representation--regressor combination; color identifies the peptide representation and marker shape identifies the regressor.}
  \label{fig:distribution}
\end{figure}

Across 60 seed-averaged representation--regressor combinations, mean (median) SPCC was 0.462 (0.466) for peptide-similarity at threshold 0.5, 0.669 (0.707) for within-target, and 0.530 (0.524) for leave-target-out evaluation (\cref{fig:landscape,fig:distribution}). Relative to within-target prediction, median losses were 0.224 and 0.160 for peptide-similarity and leave-target-out; leave-target-out exceeded peptide-similarity by 0.067. All contrasts remained significant across the evaluated configurations after Holm adjustment (two-sided Wilcoxon tests, adjusted $p<0.001$).

The top configuration also changed. ECFP-16 count fingerprints with random forest led peptide-similarity (0.545) and within-target evaluation (0.771), whereas HELM-BERT with Extra Trees led leave-target-out evaluation (0.605). Thus, the selected configuration was partition dependent; these maxima do not establish intrinsic superiority of either representation family beyond the models evaluated here.

Complete-configuration rank correlations were 0.860 for peptide-similarity versus within-target, 0.801 for peptide-similarity versus leave-target-out, and 0.795 for within-target versus leave-target-out. In the same order, representation-rank correlations fell to 0.103, 0.624, and $-0.042$, whereas regressor-rank correlations remained 0.886, 0.771, and 0.943. Partition choice therefore affected representation rankings more strongly than regressor rankings.

Tree ensembles occupied the upper part of each distribution but showed larger within-target to leave-target-out differences (\cref{fig:distribution}). Random forest, Extra Trees, MLP, and LightGBM lost 0.179, 0.177, 0.176, and 0.170 mean SPCC. SVR and ridge lost 0.067 and 0.065, although their lower within-target baselines limit interpretation of the smaller loss. Representation-level losses ranged from 0.122 (MolFormer-XL) and 0.123 (PepTune) to 0.164 (ECFP-16) and 0.165 (PepFuNN).

\subsection{Additional supervision narrows representation differences but not learner differences}

Learning curves used five pretrained peptide representations, all six regressors, and fixed validation and test sets under peptide-similarity threshold 0.5. Training sizes were 100, 300, 1,000, 3,000, 6,000, and all 8,172 available records (\cref{fig:scaling}).

Extra Trees, random forest, and LightGBM achieved the highest full-data means: 0.526, 0.518, and 0.513 SPCC. Their 100-record values were 0.186, 0.170, and 0.150. All three tree models continued to improve beyond 1,000 records (\hyperref[fig:scaling]{Figure~\ref*{fig:scaling}A,C}). From 100 records to all 8,172 records, their absolute SPCC gains were approximately 0.34--0.36. By comparison, MLP rose from 0.099 to 0.430, ridge from 0.168 to 0.424, and SVR from 0.151 to 0.403; the latter two flattened more clearly at larger sizes.

HELM-BERT had the highest representation mean at every displayed size (\hyperref[fig:scaling]{Figure~\ref*{fig:scaling}B,D}), with its clearest advantage under low-data conditions. At 100 records, SPCC was 0.210 for HELM-BERT, 0.155 for PepLand, 0.156 for PeptideCLM-2, 0.152 for PeptideCLM, and 0.099 for PepTune. Full-data values were 0.488, 0.455, 0.475, 0.462, and 0.464, reducing the best--worst gap from 0.111 to 0.033. PepTune gained the most (0.366), while HELM-BERT gained 0.279 from a stronger baseline. Among the five pretrained peptide representations, representation choice had its largest effect with limited data and became less influential as supervision increased; by contrast, substantial differences among downstream learners remained at larger training sizes.

\begin{center}
  \begin{minipage}{\linewidth}
    \centering
    \includegraphics[width=\linewidth]{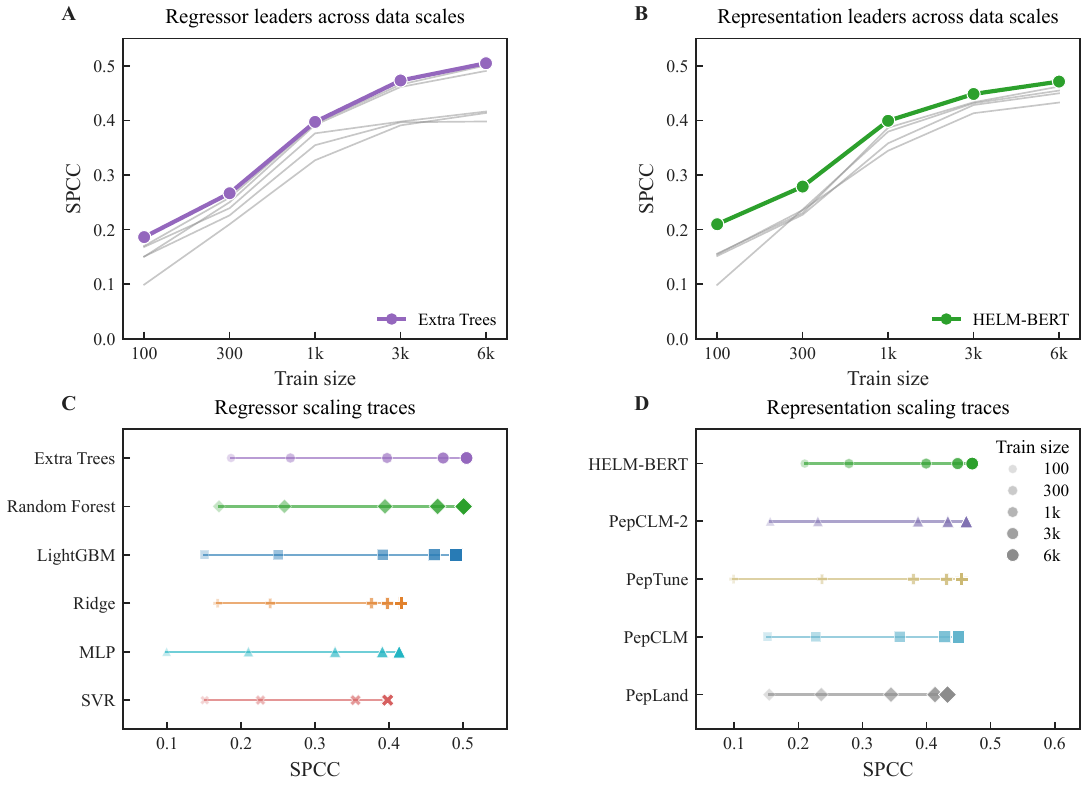}
    \captionof{figure}{Effect of training-set size on peptide--protein affinity prediction under peptide-similarity partitioning at threshold 0.5. (A) Regressor-specific curves averaged across five peptide representations, with Extra Trees highlighted as the highest-performing regressor at each shown size. (B) Representation-specific curves averaged across six regressors, with HELM-BERT highlighted as the highest-performing representation at each shown size. (C, D) Horizontal scaling traces for all regressors and peptide representations, respectively; marker size increases with training-set size. The display covers 100--6,000 records; full-data results for 8,172 records are reported in the text. Panels show descriptive means without uncertainty intervals. In panel D, PepCLM and PepCLM-2 denote PeptideCLM and PeptideCLM-2, respectively.}
    \label{fig:scaling}
  \end{minipage}
\end{center}

\subsection{PeptideCLM-2 adaptation is less influential than training-set size}

\begin{figure}[!htbp]
  \centering
  \includegraphics[width=0.70\linewidth]{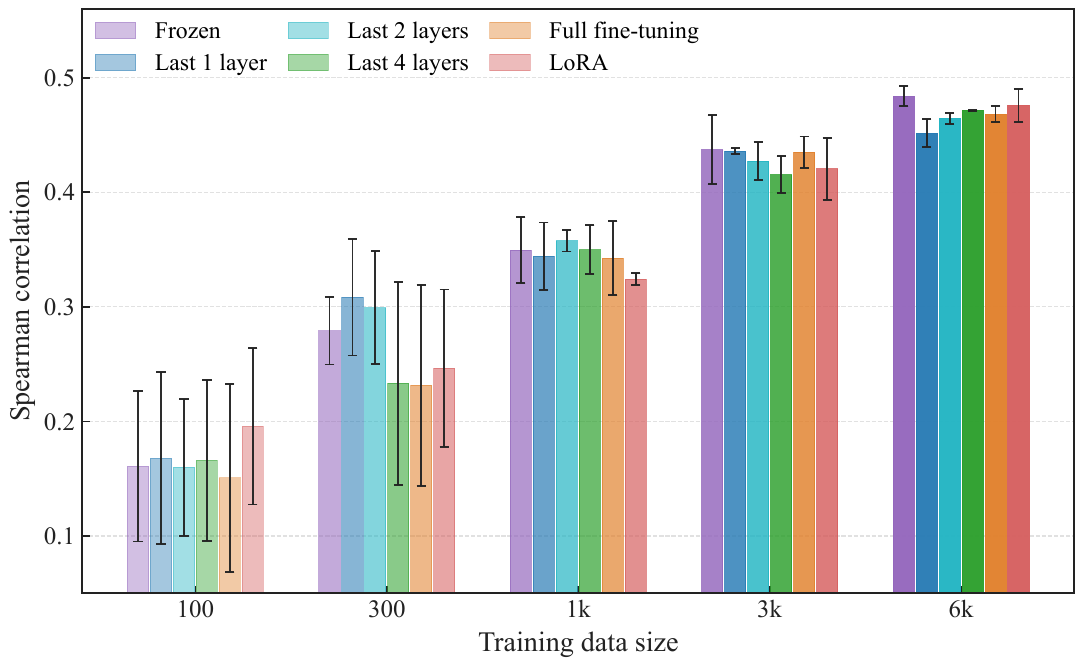}
  \caption{PeptideCLM-2 hybrid-small adaptation across training-set sizes. Test SPCC is shown for a frozen encoder, partial unfreezing of the final one, two, or four transformer blocks, full fine-tuning, and LoRA. All strategies used matched peptide-similarity partitions at threshold 0.5 and three subsampling replicates; error bars denote the between-replicate standard deviation, and opacity increases with training-set size. The display is limited to 100--6,000 records; the 8,172-record point was included in AULC but falls outside the plotted range.}
  \label{fig:finetuning}
\end{figure}

We compared frozen PeptideCLM-2 hybrid-small with Last-1, Last-2, Last-4, full fine-tuning, and LoRA across the same partition and training sizes (\cref{fig:finetuning}).

All strategies improved with more records, but differences among strategies were small relative to the training-size effect, and their relative ordering was unstable across data sizes. Between-replicate variability was greatest under low-data conditions and generally decreased as training size increased. Unfreezing one or two final blocks was competitive at several sizes; Last-4, full fine-tuning, and LoRA showed no systematic gain. At 6,000 records, mean SPCC ranged from 0.452 to 0.484 without a monotonic relation between the number of trainable blocks and performance.

The AULC comparison did not provide evidence of an overall difference among strategies (Friedman test, $\chi^2=6.62$, $p=0.251$). With only three paired subsampling replicates, this test had limited sensitivity to small differences; the numerical increase across training sizes was nevertheless much larger than the separation among adaptation strategies.

\FloatBarrier
\subsection{Explicit pairwise features provide no consistent gain}

\begin{figure}[!htbp]
  \centering
  \includegraphics[width=\linewidth]{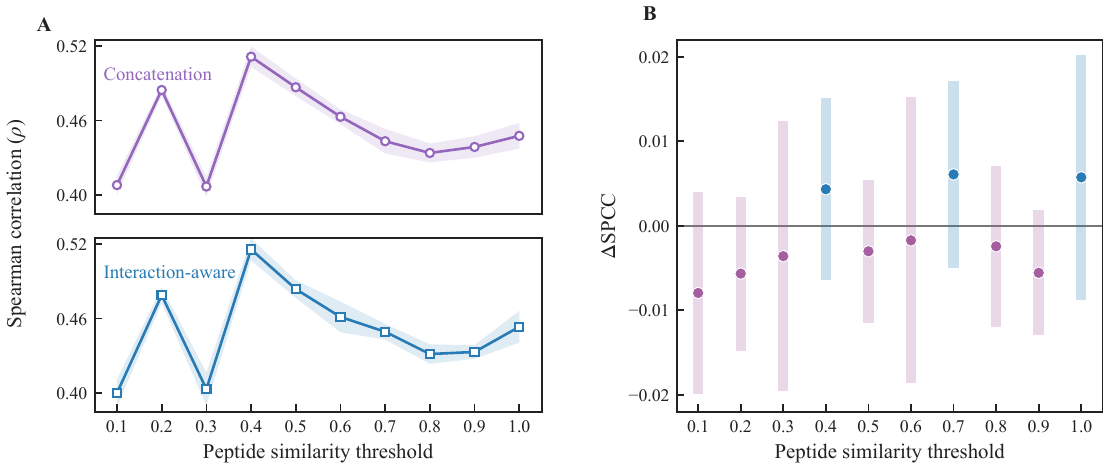}
  \caption{Comparison of peptide--protein feature-fusion rules across peptide-similarity thresholds. (A) Test SPCC for direct concatenation and interaction-aware fusion. (B) Matched difference $\Delta\SPCC=\SPCC_{\mathrm{interaction}}-\SPCC_{\mathrm{concatenation}}$; the horizontal line marks equal performance. Shaded ribbons in A and vertical intervals in B show $\pm1$ standard deviation across five matched seeds.}
  \label{fig:fusion}
\end{figure}

To test whether explicit peptide--protein interaction features improved prediction, direct concatenation was compared with interaction-aware fusion containing the element-wise product and absolute difference of projected peptide and protein features. Both used five matched seeds at peptide-similarity thresholds 0.1--1.0 (\cref{fig:fusion}).

The performance curves were highly similar throughout the threshold range: concatenation was higher at seven thresholds and interaction-aware fusion at three. The paired contrast
\begin{equation}
  \Delta\SPCC=\SPCC_{\mathrm{interaction}}-\SPCC_{\mathrm{concatenation}}
  \label{eq:fusion-difference}
\end{equation}
ranged from $-0.0079$ at threshold 0.1 to $+0.0061$ at 0.7, with no persistent advantage.

After averaging across thresholds within each seed, concatenation achieved $0.4525\pm0.0018$ SPCC versus $0.4511\pm0.0044$ for interaction-aware fusion (mean $\pm$ standard deviation across five seeds; $\Delta=-0.0014$). Interaction-aware fusion achieved the higher SPCC in 21 of the 50 matched threshold--seed comparisons. Thus, under the present model and data settings, adding element-wise products and absolute differences provided no consistent predictive gain.

\FloatBarrier
\section{Discussion}

Within this benchmark, the same 60 model configurations produced different accuracies, selected configurations, and representation rankings when the held-out entity changed. Mean SPCC was lowest under peptide-similarity partitioning, highest within targets, and intermediate when exact target sequences were excluded. This ordering should not be interpreted as an intrinsic comparison of peptide and target novelty because the partitions differ in composition and constraints. More generally, model performance cannot be attributed to the peptide encoder alone: evaluation regime, training-data scale, and downstream learner must be considered jointly.

These findings extend the chemically informed peptide evaluation of Fern{\'a}ndez-D{\'i}az et al. \cite{fernandezdiaz2025modified}. ECFP-16 count fingerprints with random forest led peptide-similarity and within-target evaluation, whereas HELM-BERT with Extra Trees led leave-target-out evaluation. Peptide-side structural novelty and target-side novelty therefore pose different generalization challenges and may favor different representation--learner combinations. The partitions map to different uses: within-target evaluation to follow-up peptide optimization or candidate ranking around an observed protein target, peptide-similarity evaluation to transfer toward new candidates chemically separated from the training peptides, and leave-target-out evaluation to protein targets absent from training. None is universally most realistic, and a single partition can conceal consequential model-selection changes \cite{walsh2021,teufel2023,tossou2024,fernandezdiaz2025good,ektefaie2024}.

Related studies predict continuous affinity, motif--domain binding, binary interactions, or residue-level binding sites \cite{giguere2013,cunningham2020,lei2021,abdin2022pepnn,li2025pepban}. Our benchmark is not a leaderboard comparison with these systems: it compares fixed representations within a unified quantitative affinity-prediction framework under controlled peptide and target shifts. Interaction-classification performance therefore cannot be assumed to imply accurate affinity ranking for an unseen target.

Data scale altered representation and learner comparisons differently. Among the five pretrained peptide representations, performance differences were largest under low-data conditions and narrowed as supervision increased, whereas tree ensembles continued to improve at larger training sizes. This agrees with the competitiveness of conventional chemical fingerprints in peptide prediction and the sensitivity of low-data chemical modeling to sampling and partition design \cite{adamczyk2026,tom2023}. It also cautions against evaluating an encoder with only one learner or at one data scale, which can misattribute learner capacity and representation--learner interactions to the representation itself. Full fine-tuning and LoRA likewise did not consistently improve the PeptideCLM-2 learning curves. Although heterogeneous binding labels may have limited reliable estimation of the added parameters, that explanation was not tested directly. Under the tested protocol, increasing supervision was associated with the more consistent numerical improvement.

The feature-fusion finding is more narrowly bounded. Element-wise products and absolute differences expose simple pairwise relations but neither model residue-level contact geometry nor allow one partner to condition the other. Their lack of a consistent gain does not exclude cross-attention, residue-pair networks, bilinear coupling, or complex-structure models that may describe peptide--protein interactions more effectively \cite{lei2021,abdin2022pepnn,li2025pepban}. It shows only that, within the present independent-encoding and MLP prediction framework, expanding the fusion representation with element-wise products and absolute differences did not reliably improve prediction.

Several limitations define the scope of inference. Combining $K_d$, $K_i$, and $IC_{50}$ on one $-\log_{10}(\mathrm{M})$ scale leaves assay and endpoint heterogeneity unmodeled. Leave-target-out excludes identical sequences but not homologous proteins, so it does not establish transfer across families or folds; within-target evaluation also emphasizes targets with enough records to populate held-out subsets. MAPc and threshold 0.5 define only one form of peptide novelty, and the merged dataset is dominated by canonical peptides. Primary results emphasize ranking by SPCC rather than calibration or endpoint-specific error, and no independent external dataset was used. Figure~4 learning-curve comparisons are descriptive and do not include uncertainty intervals. Finally, scenario tests treat model configurations as analysis units and adaptation uses only three subsampling blocks, so the reported $p$-values do not quantify uncertainty across independently sampled datasets. Sequence- and chemistry-derived features also omit resolved complex geometry; PepBDB could support a smaller structure-aware benchmark \cite{wen2019pepbdb}.

Future work should combine homology-aware target and alternative peptide-structure partitions, incorporate assay metadata, validate on independent sources, and test jointly trained partner-aware architectures. Learning curves should remain part of evaluation. The central conclusion is that representation superiority is conditional on the intended generalization regime; credible affinity benchmarks should align partitions with use, compare multiple learners, and report sensitivity to data availability.

\FloatBarrier
\bibliographystyle{unsrtnat}
\bibliography{references}

\end{document}